\documentclass[letterpaper, 10 pt, conference]{ieeeconf}  % Comment this line out if you need a4paper

\IEEEoverridecommandlockouts                              % This command is only needed if 
\usepackage{amsmath} % assumes amsmath package installed
\usepackage{amsmath,amsfonts}
\usepackage{algorithmic}
\usepackage{hyperref}
\usepackage{algorithm}
\usepackage{array}
\usepackage{mathtools}
\usepackage{textcomp}
\usepackage{stfloats}
\usepackage{url}
\usepackage{verbatim}
\usepackage{graphicx}
\usepackage{cite}
\usepackage[ruled,vlined,algo2e]{algorithm2e}

\title{\LARGE \bf
Fuzzy Ensembles of Reinforcement Learning Policies for Robotic Systems with Varied Parameters
}

\author{Abdel Gafoor Haddad$^{1}$, Mohammed B. Mohiuddin$^{1}$, Igor Boiko$^{1,2}$, and Yahya Zweiri$^{1,3,4}$ % <-this % stops a space
\thanks{This work was supported by Khalifa University grants CIRA-2020-082
and RC1-2018-KUCARS.}% <-this % stops a space
\thanks{$^{1}$A. Haddad, M. Mohiuddin, I. Boiko, and Y. Zweiri, are with the Khalifa University Center for Autonomous and Robotic Systems (KUCARS), Khalifa University, Abu Dhabi, UAE. {\tt\small \{100049699, 100059790, igor.boiko, yahya.zweiri\} @ku.ac.ae}}%
\thanks{$^{2}$I. Boiko is also with the Department of Electrical Engineering and Computer Science, Khalifa University, Abu Dhabi, UAE}%
\thanks{$^{3,4}$Y. Zweiri is also associated with the Advanced Research and Innovation Center (ARIC) and the Department of Aerospace Engineering, Khalifa University, Abu Dhabi, UAE}%
}

\begin{document}

\maketitle
\thispagestyle{empty}
\pagestyle{empty}

%%%%%%%%%%%%%%%%%%%%%%%%%%%%%%%%%%%%%%%%%%%%%%%%%%%%%%%%%%%%%%%%%%%%%%%%%%%%%%%%
\begin{abstract}

Reinforcement Learning (RL) is an emerging approach to control many dynamical systems for which classical control approaches are not applicable or insufficient. However, the resultant policies may not generalize to variations in the parameters that the system may exhibit. This paper presents a powerful yet simple algorithm in which collaboration is facilitated between RL agents that are trained independently to perform the same task but with different system parameters. The independency among agents allows the exploitation of multi-core processing to perform parallel training. Two examples are provided to demonstrate the effectiveness of the proposed technique. The main demonstration is performed on a quadrotor with slung load tracking problem in a real-time experimental setup. It is shown that integrating the developed algorithm outperforms individual policies by reducing the RMSE tracking error. The robustness of the ensemble is also verified against wind disturbance.

%Through the inverted pendulum benchmark problem, it is shown that the proposed technique reduces the number of failures significantly with just a fractional computational cost compared to domain randomization, providing a robust control that succeeds in real-world scenarios with varying and uncertain parameters. Furthermore, it is experimentally shown, on a quadrotor-slung-load system, that integrating the developed algorithm outperforms individual policies.

\end{abstract}

%%%%%%%%%%%%%%%%%%%%%%%%%%%%%%%%%%%%%%%%%%%%%%%%%%%%%%%%%%%%%%%%%%%%%%%%%%%%%%%%
\section{Introduction}

Reinforcement Learning (RL) has evolved as a potential alternative to classical control techniques in controlling dynamical systems with the assumption of having a Markov Decision Process (MDP). It possesses more advantages for systems that are highly nonlinear and stochastic since it performs the learning in an elegant trial-and-error process~\cite{sutton}.

An open challenge in RL is its transferability from simulation to reality, known as sim-to-real gap. Tackling the sim-to-real gap has been commonly done by improving the simulation model and/or improving the policy training process. In the latter category, domain randomization (DR)~\cite{random1} is a popular direction of research where a strong variability is induced in the simulator that leads to a model that can deal with such variations which would cover the real-world scenarios. While it has been argued that the benefit of the model improvement is negligible beyond a certain point, both directions impose a huge computational burden.

Another approach to improve generalization is ensemble learning. Combining multiple algorithms as an ensemble by bagging, boosting, and other methods, has been widely known in statistics and supervised machine learning leading to some of the revolutionary models such as random forests~\cite{rf}. As parallel computing gets more advanced and accessible, it becomes more rewarding to develop RL algorithms that can effectively utilize such computing power. Interesting directions as noted in~\cite{rlbookbad} include the exploitation of this computational power by having ensembles of RL algorithms. The ensemble derives its power from the diversity of its components, as the assumption is that they make mistakes on different inputs and that the majority is more likely to be correct than any individual component. Diversity usually comes from the different algorithms employed by the decision-makers, or the different inputs used to train the decision-makers.

The rest of the paper is organized as follows. in Section II the previous attempts for DR and ensembles in RL are presented and their shortcomings are highlighted. In Section III the problem formulation of the uncertain process is given, describing the possible deviations and their effect on the control problem. Section IV shows extensive simulation results along with an experimental result on a benchmark problem. Finally, in Section V we conclude with some remarks on the achieved results and the possible future directions.

\section{Related Work}
The application of domain randomization to bridge the reality gap has become particularly notable in robotics. By adjusting parameters in physics simulators, models are trained across varied environmental conditions, paving the way for sim-to-real transfer without additional real-world training for numerous RL methodologies~\cite{random1,random2,random3,random5}.

In particular, studies like~\cite{mbm23ifac,random3} succeeded in transferring trained agents from simulation to reality. These models, despite their impressive capability, showcased some limitations when applied to tasks demanding high precision. Other efforts in domain randomization focus on refining how environments are randomized. Research like~\cite{random7,random8} emphasizes structured domain randomization and adaptive domain randomization. This not only provides context to generated data but can also lead to more informative environment variations. Furthermore, the use of Bayesian optimization has been introduced for a more efficient domain randomization process~\cite{random10}. Still, challenges remain, as high-capacity models with prolonged training times are essential to ensure adaptability to all potential variations~\cite{random11}.

Ensemble learning in RL is another area of extensive exploration. Instead of relying on a single RL algorithm, researchers have utilized multiple ones to boost performance in diverse scenarios like pole balancing~\cite{ratio}, electric vehicles~\cite{electric}, and chatbots~\cite{chatbot}. Classic attempts, such as the one described in~\cite{classic_ensemb}, integrated ensemble methods with function approximators for better RL algorithm efficiency. Another study,~\cite{ensalg}, investigated blending policies originating from different value functions, resulting in approaches like majority voting and Boltzmann multiplication. Such ensemble methods were found to occasionally outpace individual RL algorithms, particularly in simpler challenges.

Techniques to further optimize ensemble learning in RL have been proposed. Rather than adhering to a fixed ensemble rule, some researchers aim for dynamic fusion methods that adapt based on a principal agent's performance~\cite{marivate2013ensemble}. The idea of a meta-algorithm for learning state or state-action values has been explored in~\cite{nnens}. This idea emphasizes the potential of committees formed by multiple agents that employ joint decisions, which, in some cases, deliver more advantageous outcomes than single agents. Another study,~\cite{selective}, advocates for a selective ensemble, whereby certain models are chosen from the overall ensemble based on their decision-making capabilities.

Recent advancements in ensemble RL techniques are spearheaded by works as~\cite{shapings}, which promote the fusion of multi-objectivization and ensemble methods. The approach utilizes heuristic information through reward shaping to create varied enriched reward signals. Combining these signals using ensemble methods potentially diminishes sample complexity. Lastly, a cutting-edge method described in~\cite{sunrise} enhances off-policy RL techniques by combining ensemble-based weighted Bellman backups with an inference method for better exploration. Despite the novelty of these ensemble methods, they often lack coordination where different policies can have conflicting decisions.

This paper addresses the aforementioned shortcomings through the following contributions:
\begin{enumerate}
    \item Developing an ensemble of RL agents that is suitable for the control of dynamic systems through fuzzy clustering.
    \item Testing the proposed algorithm on a real-time experimental setup featuring a quadrotor with a slung load.
    
    %simulations using the benchmark inverted pendulum swing-up problem as well as in real-world on a quadrotor with a slung load experimental setup.
\end{enumerate}

\section{Approach}
\subsection{Problem Setup}
RL algorithms enable an agent to interact with an environment and learn through the data generated through that interaction in discrete time steps. The environment for the RL setup is considered as an MDP. A finite MDP is composed of the states $S$, actions $A(s)$, transition function $T(s, a, s')$ that maps states and actions pair $s$, $a$ to a probability distribution over next states $s'$, and the reward $R(s, a, s')$ that computes the reward when moving from state $s$ to $s'$ as action $a$ is executed~\cite{sutton}.

In the setting of an RL problem, the agent aims to map the states to actions by learning an optimal policy. Such policy is given as the mapping that yields the largest cumulative discounted rewards from all possible states in the next steps. In the experiments, we utilize the Deep Deterministic Policy Gradient (DDPG)~\cite{ddpg}, an off-line model-free RL algorithm. On the other hand, simulations involve both DDPG and Q-learning to demonstrate the applicability of our approach to multiple RL algorithms.

The problem that we tackle can be summarized as follows: given a plant with parameters that vary, either online or offline, we would like to have a policy that can adapt and maintain the performance across the variation in these parameters without the need for additional training.

\subsection{Fuzzy Clustering}
In order to create the ensemble, a set of agents is trained for a cluster of plants where it is assumed to be sufficiently representative of the dynamics of the plants in that cluster. Fuzzy clustering has been used for generating weighted least squares-based models for dynamic systems~\cite{fuzzymodel}. It is a suitable approach for the purpose of this paper where it provides optimal clustering with a membership associated with every cluster for each of the plants. The optimization for clustering is set as follows. Let the objective function be~\cite{fuzzybook}
\begin{equation}
    J_q=\sum_{s=1}^M \sum_{j=1}^N (\mu_{sj})^q|x^s-C^j|^2
\end{equation}
subject to
\begin{equation}
    \sum_{j=1}^N \mu_{sj}=1~,~~0 < \sum_{s=1}^M \mu_{sj} < M
    \label{constraint}
\end{equation}
in which $M$ is the number of plants, $N$ is the number of clusters, $x$ is the physical parameters vector of a plant, $C$ is the center of a cluster, and $\mu_{sj}$ is the membership value of plant $s$ belonging to cluster $j$. The Lagrange cost function of the fuzzy classification is constructed as
\begin{equation}
    J(C,\mu,\lambda)=J_q-\sum_{s=1}^M \lambda_s (\sum_{j=1}^N \mu_{sj}-1).
    \label{fc3}
\end{equation}
Setting the partial derivatives with respect to the memberships and centers to zero and solving for $\mu_{sj}$ and $C^j$ yields the update equations
\begin{equation}
    \mu_{sj}=\left(\sum_{l=1}^N \frac{|x^s-C^j|}{|x^s-C^l|} ^\frac{2}{q-1} \right)^{-1}
    \label{fc1}
\end{equation}
and
\begin{equation}
    C^j=\frac{\sum_{s=1}^M \mu_{sj}^q x^s}{\sum_{s=1}^M \mu_{sj}^q}
    \label{fc2}
\end{equation}
which are evaluated iteratively until the convergence criteria are met.

\begin{figure}[htbp]
\centerline{\includegraphics[width=8cm]{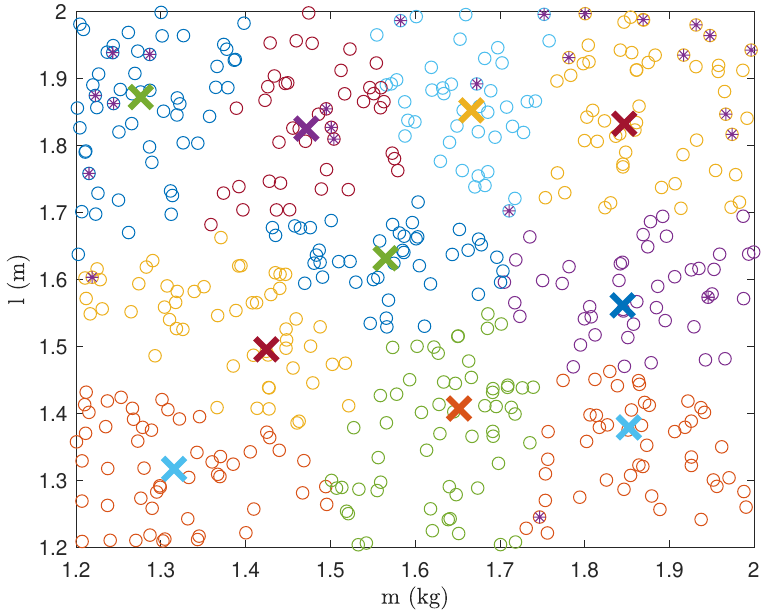}}
\caption{Illustration of the fuzzy clustering result using 10 clusters. The centers are represented by the 'X' markers whereas the sampled plants are indicated by 'o' at their respective parameters.}
\label{clusters}
\end{figure}

A sample result of clustering is shown in Fig.~\ref{clusters}. The parameter space is formed of mass and length of an inverted pendulum, and it is divided into ten clusters. The space is sampled uniformly in this example, but it can be sampled using prior knowledge about the distribution of the parameters. The centers of the clusters correspond to specific masses and lengths. RL is used to obtain ten policies, each that works successfully on the corresponding center. Circles with asterisks are for plants for which the ensemble failed to control them. It is evident that the majority of the failed plants are at the borders of the parameter space where the ensemble is expected to fail the most. In the results section, statistics show the improvement when the proposed fuzzy ensemble is deployed.

The resulting policies are fused based on weights obtained from fuzzy clustering. The Takagi-Sugeno fuzzy approach can be tailored to incorporate the fusion algorithm. Define the fuzzy set for centroids where each cluster centroid $C_i=[m_i,l_i]$ has a policy trained for it. For each $C_i$, a rule is established based on the associated policy:

IF mass $m$ is close to $m_i$ AND length $l$ is close to $l_i$, THEN policy $f_i(m,l)$. This represents the output of the policy $\pi_i$ when applied to the current state of the system.

Given $N_c$ clusters (policies), the control $u$ that is applied to plant $j$ as given by the weighted sum
\begin{equation}
    u_j=\sum_{i=1}^{N_c} \mu_{ij} \pi_i.
    \label{combo}
\end{equation}
where $\mu_{ij}$ is the membership of plant $j$ in the $i$th cluster.

It is important to note that coordination between policies is needed to ensure that the ensemble outperforms the individual policies. For this, actions from policies are compared pairwise. When the difference between two policies is below a threshold, the policies are considered to meet the agreement metric and qualify to form the ensemble. Other policies that fail to meet the metric are discarded.

By analogy from control theory, the resulting controller can be viewed as an adaptive gain scheduled control that adapts according to the operating conditions of the process which is often a practical way to compensate for variations in the process parameters or known nonlinearities~\cite{astrom}. Unlike classical control techniques such as PID where the gains are varied, it is not straightforward to vary the gains of a single agent in the context of RL. To perform scheduling in the latter, the gains (coefficients) of the linear combination of agents are varied as in~\eqref{combo}.

\subsection{Systems Models}

In this section, the mathematical models of the systems under consideration are given. For the simulations, the inverted pendulum with online identification of parameters is detailed. On the other hand, a quadrotor with slung load model is used to train RL policies and demonstrate the ensemble in real-world experiments.

\subsubsection{Inverted Pendulum}
The time-variant nonlinear model of the inverted pendulum with negligible inertia and damping is given by
\begin{equation}
    \dot{x}_1 = x_2
\end{equation}
\begin{equation}
    \dot{x}_2 = \frac{g}{l(t)}sin(x_1)+u\frac{1}{m(t)l^2(t)}
\end{equation}
where $x_1 = \Theta$ and $x_2 = \dot{\Theta}$. The pendulum length and mass are denoted by $m$ and $l$, respectively. The torque $u$ is applied directly to the pendulum pivot and $g$ is the gravitational acceleration constant.

A prior knowledge of the parameters $\psi=[l,m]^T$ is required to know the membership values in~\eqref{combo} and would allow direct execution of the control law. However, an identification algorithm is required in the case of missing or inaccurate values of the parameters. Furthermore, for time-varying parameters and to avoid interruption of operation, it is advantageous to have an algorithm that works in real-time. For this purpose, we apply Recursive Least Squares (RLS) with forgetting factor as given by~\cite{sysid}
\begin{equation}
    \hat{\psi}(t)=\hat{\psi}(t-1)+L(t)\left[y(t)-\phi^T(t)\hat{\psi}(t-1)\right]
    \label{rls}
\end{equation}
\begin{equation}
    L(t)=\frac{P(t-1)\phi(t)}{\lambda(t)+\phi^T(t)P(t-1)\phi(t)}
\end{equation}
\begin{equation}
    P(t)=\frac{1}{\lambda(t)}\left[P(t-1)-\frac{P(t-1)\phi(t)\phi^T(t)P(t-1)}{\lambda(t)+\phi^T(t)P(t-1)\phi(t)}\right]
\end{equation}
where $\hat{\psi}$ is the estimated parameters vector, $y(t)$ is the measured output, $\phi$ is the regressors vector, $\lambda$ is the forgetting factor, and $P$ is the parameters covariance matrix.

\begin{figure}[htbp]
\centerline{\includegraphics[width=6cm]{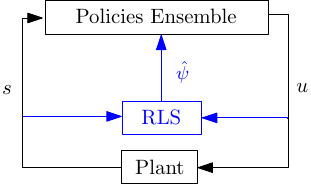}}
\caption{Block diagram of the system connection for the policy with homogeneous transformation. The adaptation loop is highlighted in blue. For the pendulum system, the state vector $s$ is composed of the two states $x_1$ and $x_2$.}
\label{homo}
\end{figure}

The execution of the ensemble of policies with the identification routine is shown in Fig.~\ref{homo}. The inputs to the RLS algorithm, which contains preprocessing and post-processing operations, are the true output $y(t)$ and the regressors $\phi(t)$. The latter is composed of the control $\phi_1(t)=u(t)$ which is the torque in this case, and $\phi_2(t)=sin(\Theta)$. The output $y(t)=\Ddot{\Theta}$ is the approximate derivative of the angular velocity which is provided by an encoder. All signals are discretized using the zero-order hold. The post-processing involves simple mathematical manipulations to retrieve the physical parameters. The coefficient of the second regressor is $g/l$ from which the length is obtained. Consequently, the mass is deduced from the coefficient of the first regressor which is $1/(ml^2)$.

\subsection{Quadrotor Slung Load System}

A quadrotor slung load system involves a quadrotor Unmanned Aerial Vehicle (UAV) transporting a load that is suspended below it using cables. This system presents unique control challenges due to the dynamic interaction between the UAV and the slung load. The swinging motion of the load can introduce instabilities to the UAV, and it is crucial to account for this in the control and design of the system.

The load's motion can be represented using a simple pendulum model, considering the tension in the cable and the forces acting on the load. The forces and moments acting on the quadrotor due to the slung load need to be incorporated into the quadrotor's dynamic equations. These forces are mainly due to the tension in the cable and any swinging motion of the load.

The dynamics of the system can be formulated by considering both the translational dynamics of the quadrotor and the angular dynamics of the load because of the assumption of the load being suspended to the center of gravity of the UAV \cite{mohiuddin2021dynamic}. The complete system is represented by a set of coupled nonlinear differential equations.

Using the notation $\mathfrak{J}=m_l L$, the equations of motion for the quadrotor in the x, y, and z directions are:

\begin{align}
    \ddot{x} &= \frac{F_x - \mathfrak{J} \Ddot{\alpha} \cos(\alpha) \cos(\beta) + \mathfrak{J} \dot{\alpha}^2 \sin(\alpha) \cos(\beta)}{m_q + m_l} \\
    \ddot{y} &= \frac{F_y - \mathfrak{J} \Ddot{\beta} \cos(\alpha) \cos(\beta) + \mathfrak{J} \dot{\beta}^2 \sin(\beta) \cos(\alpha)}{m_q + m_l} \\
    \begin{split}
    \ddot{z} &= \frac{F_z - (m_q + m_l) g}{m_q + m_l} \\
             &\quad + \frac{\mathfrak{J} \Ddot{\alpha} \sin(\alpha) \cos(\beta) + \mathfrak{J} \Ddot{\beta} \cos(\alpha) \sin(\beta)}{m_q + m_l} \\
             &\quad + \frac{\mathfrak{J} \dot{\alpha}^2 \cos(\alpha) \cos(\beta) + \mathfrak{J} \dot{\beta}^2 \cos(\alpha) \cos(\beta)}{m_q + m_l}
    \end{split}
\end{align}

The angular dynamics of the load with respect to the x and y axes are:

\begin{align}
    \begin{split}
            \ddot{\alpha} &= -\frac{F_x \cos(\alpha) + F_y \sin(\alpha) \sin(\beta)}{m_q L} \\
                 &\quad -\frac{F_z \sin(\alpha) \cos(\beta) + m_q L \dot{\beta}^2 \sin(\alpha) \cos(\alpha)}{m_q L}
    \end{split} \\
    \begin{split}
        \ddot{\beta} &= \frac{2 m_q L \dot{\alpha} \dot{\beta} \sin(\alpha) - F_y \cos(\beta) - F_z \sin(\beta)}{m_q L \cos(\alpha)} 
    \end{split}
\end{align}

\noindent where, \( m_q \) is mass of the quadrotor, \( m_l \) is mass of the slung load, \( F_{x,y,x} \) are the thrust forces along the $x, y,$ and $z$ axes, respectively, \( \alpha \) is angle of the load with respect to the x-axis, \( \beta \) is angle of the load with respect to the y-axis, \( L \) is length of the cable.

\section{Results}
\subsection{Model Training}
This section describes the hardware used for training and the time taken for each method. The algorithm parameters used for this specific study are also provided.
\subsubsection{Inverted Pendulum}
The neural networks' framework is in line with the requirements of the DDPG approach discussed in Section 2.2. It consists of deep actor and critic networks, denoted by $\pi$ and $Q$, respectively. The long-term reward is approximated through the critic value function representation based on the observations and actions. Thus, the critic is a deep neural network that is composed of two branches. The first branch takes the observations through a fully connected layer of $200$ neurons, rectified linear unit (ReLU) activation function, and another dense layer of $100$ neurons. The second branch takes the actions and processes the actions through one dense layer of $100$ neurons. The outputs of the two branches are summed and passed to a ReLU activation function to yield the critic output.

The actor network takes the current observations of the simulated system and decides the actions to take. This network starts with a fully connected layer of $200$ neurons, ReLU activation, another dense layer of $100$ neurons, and finally a hyperbolic tangent activation followed by an up-scale of the output to the torque limits. The learning rate of the actor is selected as $1\times10^{-4}$ whereas a larger learning rate of $1\times10^{-3}$ is set for the critic. The discount factor of $0.99$ and noise variance of $0.6$ are used. The simulation time is set to $20$ s and the sampling time of the agent is $0.05$ s, leading to a $400$ steps per simulation. These values of hyper-parameters were obtained by tuning them to achieve the required swing-up task.

The reward at every time step is given by
\begin{equation}
    r_t=-(\Theta_t^2+0.1\dot{\Theta}_t^2+0.0001u_{t-1}^2).
\end{equation}
Through this reward, the RL algorithm receives feedback to know about goodness of the actions that have been executed. In a realistic scenario, the measurements of the angle and the angular rate are accessible through encoders.

The parameters of the pendulum are set to $m=1$~kg and $l=9.81$ m (value of $g$). Furthermore, the input torque is saturated to $[-30,30]$ Nm during the training such that the pendulum cannot swing up without oscillating.

\subsubsection{Quadrotor with Slung Load}
The network structure used for the quadrotor is similar to the one used for the pendulum except that the capacity of the layers in the network is reduced to $200$ and $100$ neurons respectively. This simpler network was capable of learning the velocity control of the quadrotor with slung load.

\subsection{Simulation Results}
\subsubsection{Inverted Pendulum}
This section shows the simulations of the inverted pendulum. The pendulum starts at the downward position which corresponds to $\pi$ rad and is swung up to the balanced position of $0$ rad. Also, the angle is wrapped to the interval $[-\pi,\pi]$. The action which is the torque, is continuous and its absolute value is saturated at $50$ Nm.

\begin{figure}[htbp]
\centerline{\includegraphics[width=8.5cm]{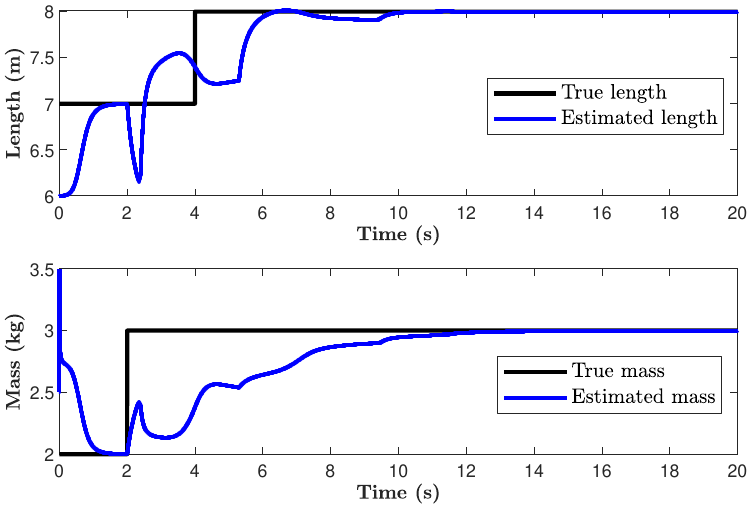}}
\caption{True values of the length and mass along with their on-line estimations using RLS.}
\label{parameters}
\end{figure}

The proposed method is tested with the parameters of the system are unknown. Fig.~\ref{parameters} shows the estimations of the length and the mass using RLS along with their true values. The initial estimate of the length is $6$ m while the true value is $7$ m. Furthermore, the length changes to $8$ m in a step at $t=4$ s. For the mass, the initial estimate is $2.5$ kg whereas the true initial value is $2$ kg. The mass steps up to $3$ kg at $t=2$ s. The estimations are performed in real-time and converge to the true values while the pendulum is swinging up. A forgetting factor of $0.998$ is used to discount the older data and facilitate the convergence.

\begin{figure}[htbp]
\centerline{\includegraphics[width=8.5cm]{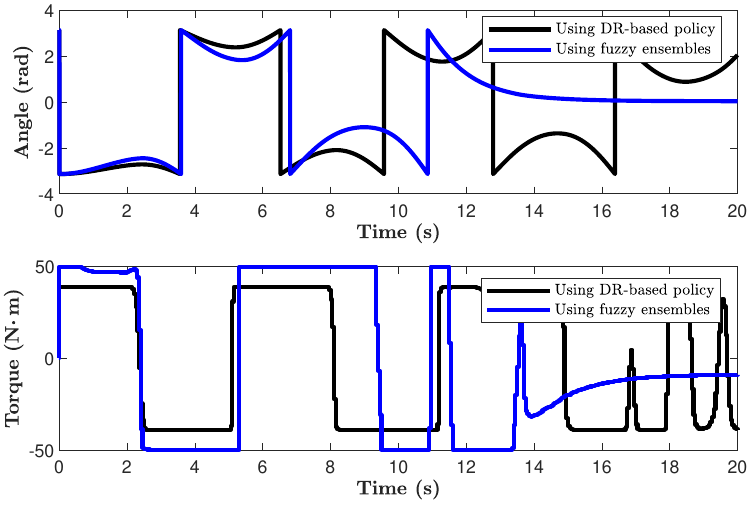}}
\caption{System response and control torque under variation of the length and mass.}
\label{resp_var}
\end{figure}

The training of the RL agent with DR was performed by varying the length and the mass independently over the intervals $[6,8]$ m and $[0.5,2.5]$ kg through a uniform distribution. Although the parameters are varied within the randomization interval used during training, DR-based policy failed to swing up and stabilize the pendulum at the upward position as shown in Fig.~\ref{resp_var}. It can be seen that the quick estimation of the parameters using RLS leads to a successful swing-up using the ensemble. The robustness inherited by the RLS adaptation loop allows the proposed method to be used with systems with uncertainties in their parameters.

\begin{figure}[htbp]
\centerline{\includegraphics[width=8cm]{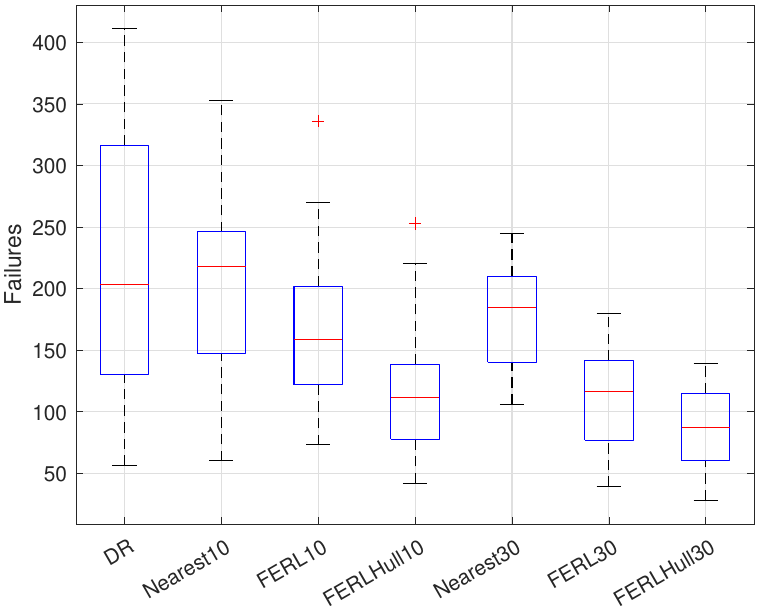}}
\caption{Statistics of the failures for the inverted pendulum. The results are obtained through 100 trials to account for the learning stochasticity.}
\label{box2}
\end{figure}

For more quantitative analysis, improvements in the success rate of swinging up the pendulum are reported in the statistics shown in Fig.~\ref{box2} in a similar manner to other RL literature~\cite{assessing}. The box plot shows failure rate statistics for 1600 pendulums of different parameters when 10 clusters are formed. The Fuzzy Ensemble of Reinforcement Learning (FERL) policy is compared to DR as well as the case when the nearest policy is executed. Furthermore, the case when only the policies forming a hull around the test plant (FERLHull) are used in the fuzzy fusion is shown to lead to lower failures. Another set of results is also reported for the case when 30 clusters are formed. This indicates that the failure rate decreases with a higher number of trained policies. This is expected since in the extreme case, one policy is trained for each possible plant. However, this increase comes at the cost of computational complexity due to training more agents. The code to reproduce the results is open-sourced on Matlab Central\footnote{mathworks.com/matlabcentral/fileexchange/135447-rl-ensemble}.

\subsection{Real-World Experiments}
The real-world experiments are performed on a quadrotor with slung load. We show the performance in tracking a 3D trajectory in comparison with the plain RL-based policies.

The experimental setup is based on the Quanser QDrone. It weighs $M=1264$ g with a thrust-to-weight ratio of 1.9 allowing it to carry $m_{max}=300$ g with a reasonable maneuvering ability. The on-board processing is performed on the Intel Aero Compute unit which is available off the shelf. The Madgwick filter~\cite{mad} estimates the attitude of the QDrone using the BMI160 inertial measurement unit mounted on board. The position and yaw measurements of both the drone and the load is provided by the OptiTrack motion capture system at 250 Hz. Further details about the UAV parameters can be found in~\cite{dnn}. The angles that the load makes with respect to the drone are deduced from their relative positions. The payload is manufactured from mild steel to reduce its size and minimize its interaction with the propellers. The load weighs $m_b=62$ g and it is equipped with three markers for localization.

First, experiments are performed using individual RL policies that have been trained to track the reference trajectory when the load has a cable length of $0.5$ m and $1$ m, respectively. However, the cable length in the actual test setup is $L=0.75$ m as shown in the accompanied video\footnote{\url{https://youtu.be/2PH8WVGFTt0}}. After running the individual policies, a fuzzy ensemble of policies is executed to study the effectiveness of the proposed technique. Finally, wind disturbance is applied towards the end of the flight to test the disturbance rejection of the ensemble. The wind profile follows~\cite{prav}. 

\begin{figure}[htbp]
\centerline{\includegraphics[width=8cm]{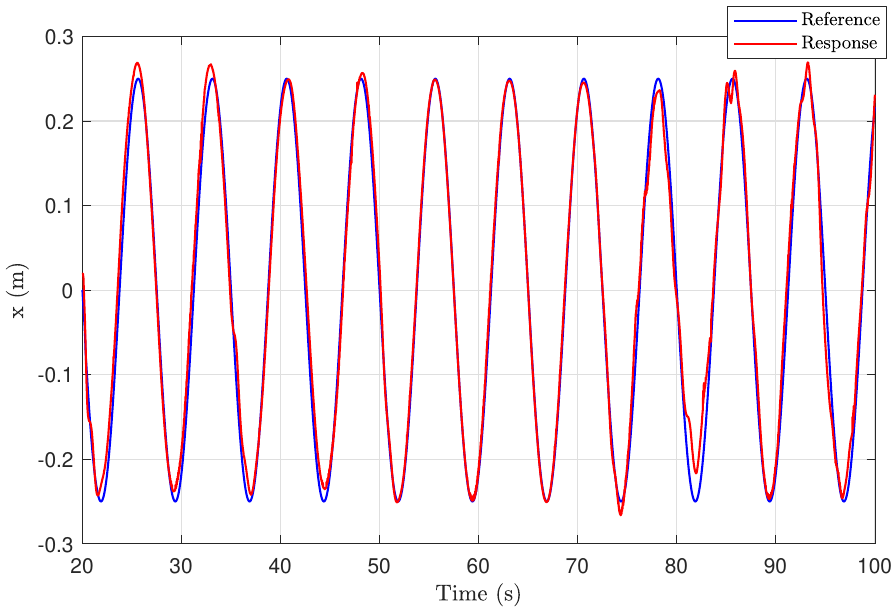}}
\caption{Quadrotor x-position response.}
\label{exp_resp1}
\end{figure}

The trajectory tracking response is shown in Fig.~\ref{exp_resp1} for one of the directions. At $t=20$ s, the agent trained for $L=0.5$ m is applied. At $t=35$ s, the control is switched to the agent trained for $L=1$ m. At $t=50$ s, the controller switches to the fuzzy ensemble. At $t=65$ s, the wind disturbance is applied. It can be noticed that from $t=20$ s to $t=50$ s, an error exist in the tracking, which is more noticeable near the peaks. After the execution of the ensembled policy at $t=65$ s, the error becomes negligible. The application of wind disturbance introduced an error again, but the ensemble shows a disturbance rejection capability, indicating learning an integral action.

\begin{figure}[htbp]
\centerline{\includegraphics[width=8cm]{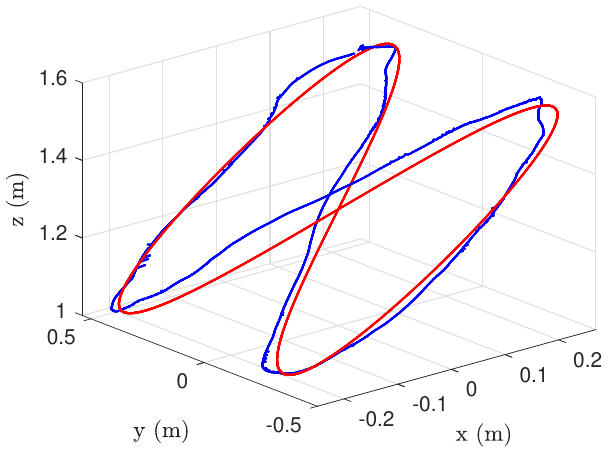}}
\caption{Response.}
\label{exp_resp2}
\end{figure}

The 3D plot of the last cycle when the wind is applied is shown in Fig.~\ref{exp_resp2}. A higher error is noticed in the y-position which could be due to the way the slung load is connected to the quadrotor base, which allows higher oscillations in that direction.

A summary of the root-mean-square error is presented in Table 1. A significant reduction in the RMSE can be noticed in the $x$ and $z$ directions whereas the reduction is minor in $y$. This can be attributed to the nature of the ensemble process in which more success is expected for cases where the agreement between the individual policies is higher. During the wind disturbance of the same window length, $65\leq t \leq 80$, the ensemble managed to limit the total RMSE to $0.0524$ m.
\begin{table}[h]
\caption{Tracking RMSE}
\centering
\begin{tabular}{ |p{0.5cm}||p{1.55cm}|p{1.55cm}|p{1.55cm}|  }
 \hline
 &$20\leq t \leq 35$ &$35\leq t \leq 50$&$50\leq t \leq 65$\\
 \hline
$x$ &0.0256 &0.0177 &\textbf{0.0066}\\
$y$ &0.0363 &0.0349 &\textbf{0.0325}\\
$z$ &0.0208 &0.0212 &\textbf{0.0086}\\
3D &0.0490 &0.0445 &\textbf{0.0343}\\
 \hline
\end{tabular}
\label{3class}
\end{table}

\section{Conclusion}
In this paper, a new ensemble method for RL policies is proposed and investigated. The simulation results show that the developed method contributes to reducing the number of failures in the inverted pendulum problem, and outperforms DR for the case of time-variant pendulum. The experimental results for the quadrotor with slung load support the conclusion that the ensemble outperforms individual policies. These results indicate that ensemble learning for RL is a promising direction for the control of real-world robotic systems.

Future work may include continuous and model-based RL algorithms, which can be very useful in reducing the number of experiences. Moreover, we anticipate that an ensemble composed of a variety of RL methods, exploiting their complementary features, may provide better results than the ensemble of a single method.

\bibliographystyle{IEEEtran}
\bibliography{ref}

\end{document}